\pdfoutput=1

\documentclass[11pt]{article}

\usepackage[final]{acl}
\usepackage{times}
\usepackage{latexsym}
\usepackage{booktabs}
\usepackage[T1]{fontenc}
\usepackage[utf8]{inputenc}
\usepackage{microtype}
\usepackage{inconsolata}
\usepackage{graphicx, amsmath}
\usepackage{enumitem}
\newcommand{\gptfouro}{\textsc{GPT-4o}}
\newcommand{\gemini}{\textsc{Gemini-1.5-Flash}}
\newcommand{\llamathree}{\textsc{LLaMA-3.3-70B-Instruct}}
\newcommand{\llamathreeone}{\textsc{LLaMA-3.1-8B-Instruct}}

\newcommand{\mathset}{\textsc{MATH}}
\newcommand{\gsmset}{\textsc{GSM8K}}

\title{Not the Example, but the Process: \\ How Self-Generated Examples Enhance LLM Reasoning}

\author{
  Daehoon Gwak$^{1}$ \quad
  Minseo Jung$^{2}$ \quad
  Junwoo Park$^{1}$ \quad
  Minho Park$^{1}$ \\
  \textbf{ChaeHun Park$^{1}$ \quad
  Junha Hyung$^{1}$ \quad
  Jaegul Choo$^{1}$} \\\\
  $^{1}$KAIST AI \\
  $^{2}$Applied Artificial Intelligence, Sungkyunkwan University \\
  \texttt{\{daehoon.gwak, jchoo\}@kaist.ac.kr} \quad
  \texttt{jms020123@g.skku.edu}
}

\begin{document}
\maketitle
\begin{abstract}
Recent studies have shown that Large Language Models (LLMs) can improve their reasoning performance through self-generated few-shot examples, achieving results comparable to manually curated in-context examples. 
However, the underlying mechanism behind these gains remains unclear, making it hard to decide when and how to apply the technique effectively.
In this work, we argue that \textbf{the key benefit arises not from the generated examples themselves but from the act of creating them}.
To validate this, on reasoning-intensive tasks across diverse LLM architectures, we systematically evaluate three prompting strategies for in-context learning: (1) Zero-shot prompting; (2) Integrated prompting, where LLMs create and solve problems within a single, unified prompt; and (3) Decoupled prompting, where self-generated examples are reused as in-context examples, but the context of their creation itself is excluded. 
We conduct experiments across five widely used model architectures, demonstrating that Integrated prompting consistently outperforms both Zero-shot and Decoupled prompting. 
In contrast, Decoupled prompting offers only marginal gains over Zero-shot.
Further, for a more in-depth analysis, we conduct an attention analysis and observe significant differences in attention patterns between Integrated and Decoupled prompting.
These findings suggest that the advantage of self-generation prompting comes from the process of problem creation, not the examples themselves, providing valuable insights for designing more effective prompting strategies.
\end{abstract}

\section{Introduction}
Large Language Models (LLMs) have demonstrated remarkable capabilities in tackling complex reasoning tasks, such as mathematical problem solving \citep{brown2020language,Rae2021ScalingLM,chowdhery2023palm,achiam2023gpt}.
In particular, few-shot in-context learning \citep{brown2020language} has emerged as a powerful approach for enhancing LLM reasoning, along with advances in prompting techniques \citep{wei2022chain,kojima2022zerocot}.
By leveraging a small number of examples as guidance, these approaches allow models to reason systematically and effectively, without the need for extensive task-specific fine-tuning. 
However, these approaches often rely on labeled exemplars, either manually curated or retrieved, to serve as reasoning guides. 
This dependency introduces challenges such as the cost of annotation, the need for task-specific exemplars, and limitations in adaptability to unseen tasks or domains.

To address these challenges, several studies have explored self-generation approaches, where LLMs generate their own few-shot examples \citep{Kim2022SelfGeneratedIL,li2024are,yasunaga2024large}. 
These approaches involve LLMs generating analogous examples or high-level tutorials to replace manually curated few-shot examples before solving the given task. 
They have shown that these self-generation prompts can adapt to diverse reasoning tasks and complexity levels, while achieving performance levels comparable to manually curated examples. 
This highlights their potential to reduce reliance on labeled exemplars, especially in scenarios where task-specific data is scarce or expensive to obtain. 
However, why this technique works is still unclear, which makes it hard to know when and how to apply it effectively. 
Therefore, in this work, we examine how self-generation prompting affects LLM reasoning.


\begin{figure*}[t] 
    \setlength{\abovecaptionskip}{12pt} 
    \centering
    \includegraphics[width=1.0\textwidth]{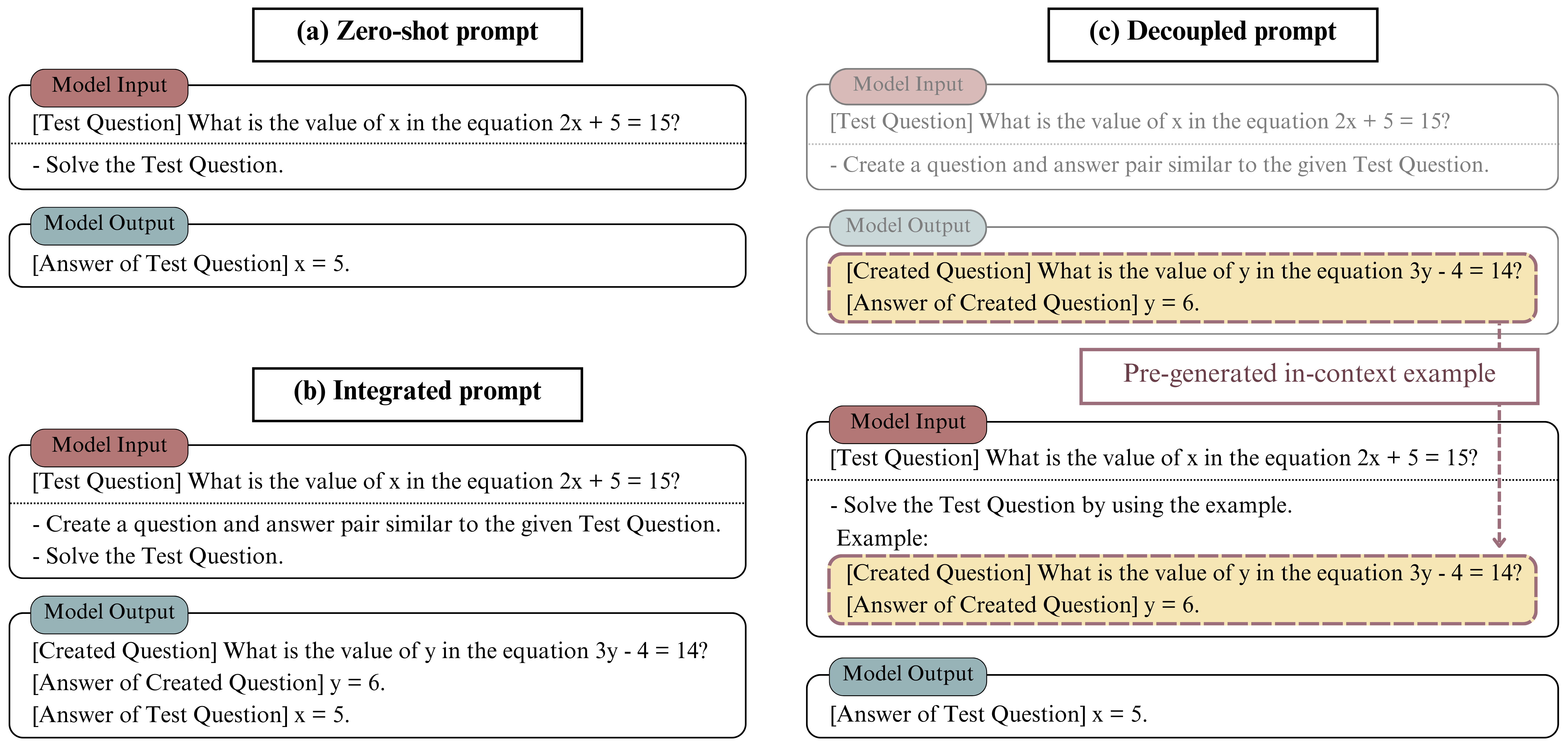}
    \caption{
    \textbf{Overview of the prompting strategies evaluated in this work.}
    Each strategy is depicted under the same test question to illustrate how the interaction between the user and assistant differs. 
    \textbf{(a) Zero-shot} involves no context; \textbf{(b) Integrated} combines problem creation and solving in a single prompt; 
    \textbf{(c) Decoupled} leverages self-generated examples as in-context exemplars without including the creation process in the conversational history.
    }
    \label{fig:concept}
    \vspace{-5mm}
\end{figure*}

To better understand the role of self-generation prompting in enhancing reasoning performance, we conduct systematic experiments across multiple prompting setups. 
As illustrated in Figure~\ref{fig:concept}, we evaluate three distinct setups, highlighting how problem creation and solving are structured differently across strategies: 
(1) solving the test question directly without additional context (\textbf{Zero-shot}), (2) generating a related problem and then solving the test question within a single prompt (\textbf{Integrated}), and (3) separating the generation and solving steps, where self-generated examples are leveraged as in-context exemplars to solve the test question (\textbf{Decoupled}). 
By comparing these setups, we can isolate the role of problem creation in LLM reasoning and analyze its impact separately.
A detailed description of these prompting strategies is provided in Appendix~\ref{sec:method}.

We evaluate these strategies on two prominent reasoning datasets, \mathset \citep{hendrycksmath2021} and \gsmset \citep{cobbe2021gsm8k}, using five representative models: \textsc{GPT-4.1}, \gptfouro \citep{achiam2023gpt}, \gemini \citep{Reid2024Gemini1U}, \llamathree, and \llamathreeone \citep{Dubey2024TheL3}.
Our experiments reveal the following key findings:
\begin{enumerate}
    \item \textbf{Integrated self-generation prompts outperform other setups.} For instance, on \mathset, \textsc{GPT-4o} achieves 37.86\% with Integrated prompting, compared to 33.33\% for Zero-shot and 32.65\% for Decoupled.
    Similarly, across \textsc{GPT-4.1}, \textsc{Gemini-1.5-Flash}, \textsc{LLaMA-3.3-70B-Instruct}, and \llamathreeone, Integrated prompting consistently yields higher accuracy on \textsc{MATH} and \textsc{GSM8K} than Zero-shot or Decoupled.
    \item \textbf{The generation process, not the generated problems, drives improvement.} Decoupled prompting, which uses the self-generated examples as Integrated prompting but without combining the problem-creation process, performs similarly to Zero-shot. This suggests the key factor is the process of creating problems, rather than the examples themselves.
    \item \textbf{Attention pattern analysis further supports the advantage of Integrated prompting.}
    Using the open-source \textsc{LLaMA-3.1-8B-Instruct}, we compared attention patterns under Integrated and Decoupled setups. Our analysis reveals significant differences in how attention is allocated, depending on whether the problem-creation occurs within the same prompt. This offers deeper insights into why Integrated prompting is more effective.    
\end{enumerate}

All these results highlight that the key factor in reasoning enhancement is not the generated examples, but the problem-creation process itself.
This emphasizes the critical role of the problem-creation within the reasoning workflow, providing practical insights for designing more effective prompting strategies for complex reasoning tasks.

\section{Related Work}

\subsection{Prompting Techniques and Self-Generated Examples}

Prompt design plays a fundamental role in guiding large language models (LLMs) to solve complex tasks. The seminal work by \citet{brown2020language} established that few-shot in-context learning enables models to adapt to a variety of tasks with only a handful of examples. Building on this insight, \citet{wei2022chain} demonstrated that chain-of-thought prompting—by providing step-by-step exemplars—can further enhance reasoning performance. Moreover, \citet{kojima2022large} found that even minimal instructions (e.g., appending “think step by step” to a query) can trigger reasoning abilities in a zero-shot setting.

More recent studies have explored automatic prompt and example generation to reduce reliance on handcrafted data. For example, \citet{min2022rethinking} and \citet{Kim2022SelfGeneratedIL} propose methods in which models generate their own demonstrations. In addition, \citet{yasunaga2024large} and \citet{li2024are} show that self-generated analogical examples can serve as effective in-context examples, often matching or even surpassing human-crafted ones. These advances suggest that self-generated examples can bridge the gap between static prompting and dynamic reasoning, offering a more flexible framework for guiding LLM reasoning performance.

\subsection{Self-Generated Reasoning}

Beyond generating examples, another line of work focuses on enabling LLMs to produce their own internal reasoning processes—often referred to as self-generated reasoning or chain-of-thought generation. This approach allows models to explicitly articulate intermediate steps leading to a final answer. For instance, \citet{zelikman2022star} introduce the STaR method, which leverages self-generated chain-of-thought data in a bootstrapping manner to iteratively refine the model's reasoning. Similarly, \citet{peng2024regenesis} propose a multi-stage process where the model initially produces abstract reasoning steps before elaborating them into complete solutions. Techniques such as self-consistency decoding \citep{wang2023self} and self-refinement \citep{madaan2023selfrefine} further empower models to generate multiple candidate reasoning paths and subsequently select or improve upon the best outcome. Overall, these methods underscore that enabling LLMs to autonomously produce detailed reasoning content is crucial for achieving enhanced performance.

\subsection{Iterative Self-Refinement and Verification}

Recent work also explores iterative refinement and verification methods, where an LLM repeatedly checks and revises its own output to improve quality. \citet{madaan2023selfrefine} propose a self-refinement loop that detects and corrects mistakes in the model’s chain-of-thought, while \citet{shinn2023reflexion} enable models to revisit earlier solution steps upon finding inconsistencies. These approaches can be integrated with self-consistency decoding \citep{wang2023self} to generate and compare multiple reasoning paths. Recent studies have also investigated structured verification prompts \citep{yao2023tree}, as well as external tool usage \citep{mialon2023augmented}, for further reasoning performance gains. By automatically identifying and repairing logical gaps, iterative self-refinement and verification offer a promising framework for enhancing LLM reasoning without heavy reliance on human intervention.

\section{Prompting Strategies and Empirical Evaluation}

In this section, we evaluate the effectiveness of self-generation prompting methods on two prominent mathematical reasoning datasets: \textsc{MATH} and \textsc{GSM8K}. Mathematical reasoning tasks are particularly widely used to analyze the reasoning capabilities of LLMs, as they require precise logical inference and structured problem-solving steps \citep{saxton2018analysing,hendrycksmath2021,cobbe2021gsm8k,zelikman2022star,lewkowycz2022solving}. We compare three prompting strategies, focusing on their impact on reasoning capabilities and the role of the self-generation process.

\begin{table*}[t]
\centering
\resizebox{0.85\textwidth}{!}{
\begin{tabular}{l@{\hspace{4pt}}ccc@{\hspace{4pt}}|@{\hspace{4pt}}ccc}
\toprule
& \multicolumn{3}{c}{\textbf{GSM8K}} & \multicolumn{3}{c}{\textbf{MATH}} \\
\cmidrule(l){2-4} \cmidrule(r){5-7} 
\textbf{Model} & \textbf{Zero-shot} & \textbf{Integrated} & \textbf{Decoupled} & \textbf{Zero-shot} & \textbf{Integrated} & \textbf{Decoupled} \\
\midrule
\textsc{GPT-4.1} & 48.825 & \textbf{57.922} & 49.052 & 33.265 & \textbf{39.425} & 37.449 \\
\gptfouro        & 49.886 & \textbf{54.587} & 50.493 & 33.333 & \textbf{37.860} & 32.649 \\
\gemini & 36.012 & \textbf{41.168} & 36.543 & 34.362 & \textbf{38.398} & 36.345 \\
\textsc{LLaMA-3.3-70B}   & 32.672 & \textbf{35.191} & 33.588 & 25.257 & \textbf{28.337} & 26.694 \\
\textsc{LLaMA-3.1-8B}   & 12.061 & \textbf{16.488} & 11.603 & 6.982 & \textbf{12.936} & 9.856 \\
\bottomrule
\end{tabular}
}
\caption{
\textbf{Accuracy (\%) on the \gsmset (left) and \mathset (right) datasets.}
Results are shown for \textsc{GPT-4.1}, \gptfouro, \gemini, \llamathree, and \llamathreeone under three prompting methods (Zero-shot, Integrated, Decoupled). 
Integrated prompting consistently outperforms both Zero-shot and Decoupled prompting across all models and datasets, while Decoupled prompting shows only marginal gains or even drops below Zero-shot in some cases.}
\label{tab:results_combined}
\vspace{-5mm}
\end{table*}

\subsection{Setup}

\paragraph{Datasets.}
We conduct experiments on two datasets, widely used as benchmarks for evaluating the reasoning capabilities of LLMs:

(1) \textbf{MATH} \citep{hendrycksmath2021}: A challenging dataset containing 12,500 competition-level mathematical problems, designed to test advanced problem-solving skills. We evaluate on its subset of 5,000 problems spanning seven mathematical subjects, including Algebra, Geometry, and Probability, and five levels of difficulty, ranging from easy (Level 1) to extremely challenging (Level 5). 

(2) \textbf{GSM8K} \citep{cobbe2021gsm8k}: A dataset of 8,000 grade-school math word problems, commonly used to assess reasoning abilities in simpler yet diverse scenarios. We use its standard test set of 1,319 problems for evaluation.

\paragraph{Metrics.} Accuracy is used as the primary evaluation metric, calculated as the percentage of problems where the model’s final answer matches the ground truth. To ensure stable and deterministic outputs, we fix the model’s generation settings to \texttt{temperature} = 0.0 and \texttt{top\_p} = 1.0 throughout all experiments.

\paragraph{Models.} We conduct experiments using five language models: \textsc{GPT-4.1}, \gptfouro, \gemini, \llamathree, and \llamathreeone. Each model is evaluated on both \textsc{MATH} and \textsc{GSM8K} to provide a comprehensive comparison of prompting strategies. In addition, we perform detailed attention-pattern analyses on the open-source \textsc{LLaMA-3.1-8B-Instruct} model under Integrated and Decoupled prompting settings.

\subsection{Prompting Methods}

We evaluate three distinct prompting strategies to investigate the impact of self-generation and compare them with Zero-shot baseline:

\paragraph{Zero-shot} The model directly solves the test question without any additional context or prior examples.

\paragraph{Integrated} The model generates a related question and its answer, then directly solves the test question, all within a single prompt. This setup tightly integrates problem creation and solving into a unified process.

\paragraph{Decoupled} The generated question and answer are presented as in-context examples for solving the test question, but the problem creation process itself is excluded from the conversational history. This setup simulates scenarios where examples are prepared externally, rather than integrated dynamically into the reasoning process. In other words, while the model can leverage the generated examples as guidance, it lacks direct access to the problem creation process itself.

Figure~\ref{fig:concept} provides an overview of the three prompting strategies. Each strategy is depicted under the same test question to illustrate how user-assistant interactions differ across methods. A more detailed description of these prompting strategies is provided in Appendix~\ref{sec:method}, and the full prompts used in our experiments can be found in Appendix~\ref{sec:prompt_examples}.

\begin{table}[t]
\centering
\begin{tabular}{lccc}
\toprule
& \textbf{1-shot} & \textbf{2-shot} & \textbf{4-shot} \\
\midrule
\textbf{Integrated} & \textbf{35.191} & \textbf{38.931} & \textbf{38.244} \\
\textbf{Decoupled}  & 33.588 & 31.908 & 32.443 \\
\bottomrule
\end{tabular}
\caption{
\textbf{Multi-shot results with Accuracy(\%) on \textbf{GSM8K}.}
Using \llamathree, we compare Integrated vs.\ Decoupled 
under 1, 2, and 4-shot conditions. Even as the number of provided examples changes, 
Integrated continues to outperform Decoupled.
}
\vspace{-5mm}
\label{tab:multi_shot_llama33_gsm8k}
\end{table}

\subsection{Results}

Table~\ref{tab:results_combined} presents the accuracy of three prompting methods---Zero-shot, Integrated, and Decoupled---across five models (\textsc{GPT-4.1}, \textsc{GPT-4o}, \textsc{Gemini-1.5-Flash}, \textsc{LLaMA-3.3-70B-Instruct}, and \textsc{LLaMA-3.1-8B-Instruct}) on both \textsc{MATH} and \gsmset. We observe a consistent pattern:

\paragraph{Integrated Outperforms Other Prompting Methods.}
Across all model--dataset pairs, Integrated prompting achieves the highest accuracy. For instance, on \gsmset, \textsc{GPT-4o} reaches 54.59\% with Integrated prompting, surpassing both Zero-shot (49.89\%) and Decoupled (50.49\%). A similar pattern is observed on \mathset, where \textsc{GPT-4o} achieves 37.86\% with Integrated, outperforming Zero-shot at 33.33\% and Decoupled at 32.65\%. Similar patterns emerge for \gemini, \llamathree, and \llamathreeone, suggesting that Integrated prompting offers stronger results regardless of model size or architecture.



\paragraph{Decoupled Shows Marginal or Inconsistent Improvements.}
Although Decoupled uses the self-generated examples as Integrated, its performance remains close to or even below Zero-shot in certain cases. For example, on \gsmset with \llamathreeone, Decoupled (11.60\%) slightly underperforms Zero-shot (12.06\%). Meanwhile, Integrated achieves 16.49\% on this setup. All these outcomes indicate that the main benefit does not simply come from using self-generated examples but from the process of generating problems within a single prompt.



Overall, these results support our main claim: the key factor driving performance gains in self-generation prompting strategies is the combined creation-and-solving process, rather than the final created examples alone. Additional results and analysis, including attention-pattern studies on \llamathreeone, can be found in the following sections.

\subsection{Analysis}
\label{subsec:analysis}
In this subsection, we provide additional experiments and in-depth analysis to better understand why Integrated prompting consistently outperforms Decoupled. We begin by exploring whether the same trend holds when more examples are given. We then examine the internal attention patterns of an open-source model under Integrated and Decoupled prompting settings.

\begin{figure}[ht]       
    \centering           
    \includegraphics[width=\columnwidth]{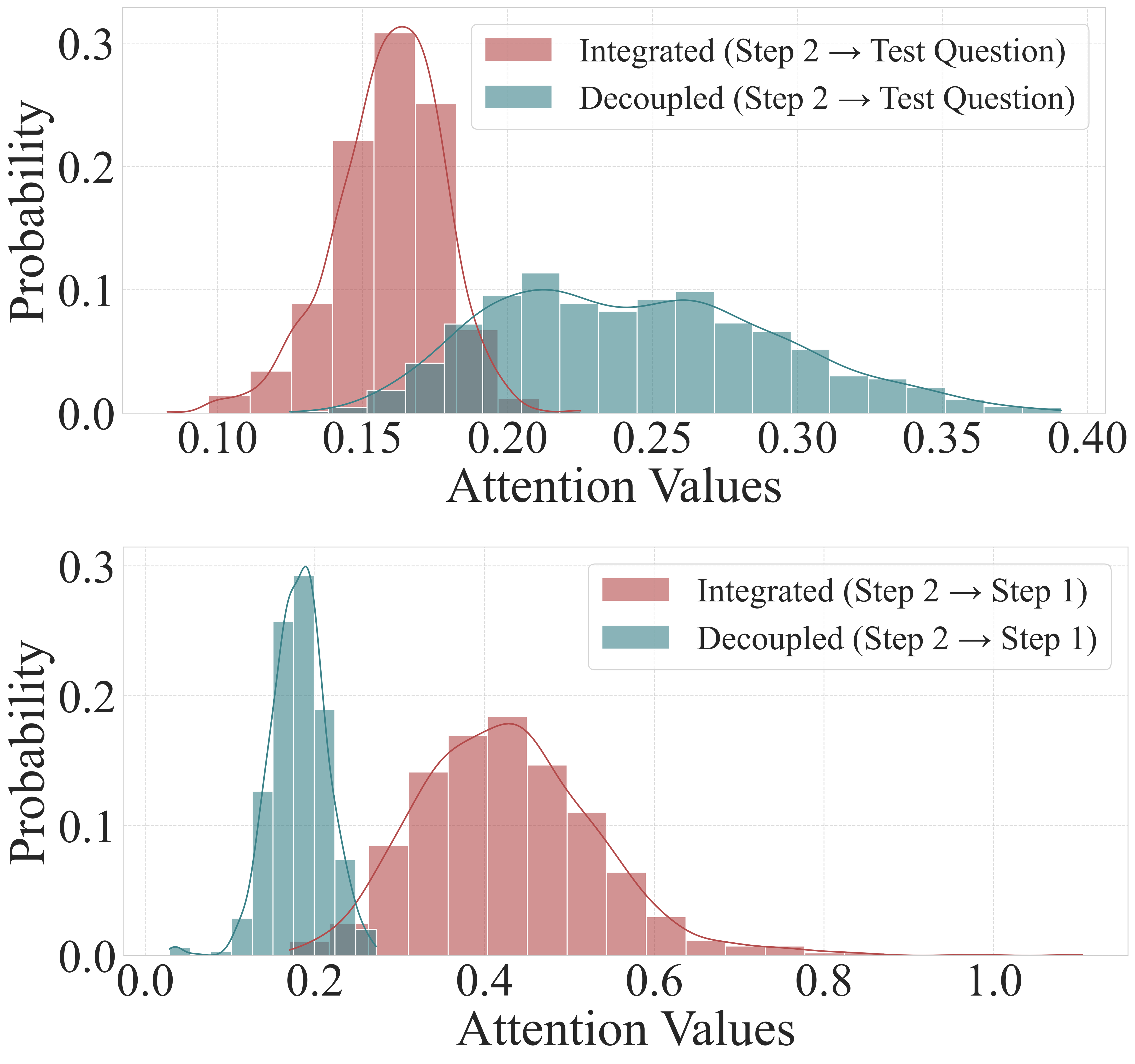}
    \vspace{-5mm}
    \caption{\textbf{Distribution of attention during test-question solving.}
    The upper histogram shows attention to test-question tokens, with \textit{Decoupled} devoting significantly more attention.
    The lower histogram shows attention to self-generated example tokens, with \textit{Integrated} exhibiting significantly higher attention (\(p\!<\!10^{-10}\), paired t-test).}
    \label{fig:attention_distribution_test}
    \vspace{-5mm}
\end{figure}

\begin{figure}[htbp]          
    \vspace{-3mm}             
    \includegraphics[width=\columnwidth]{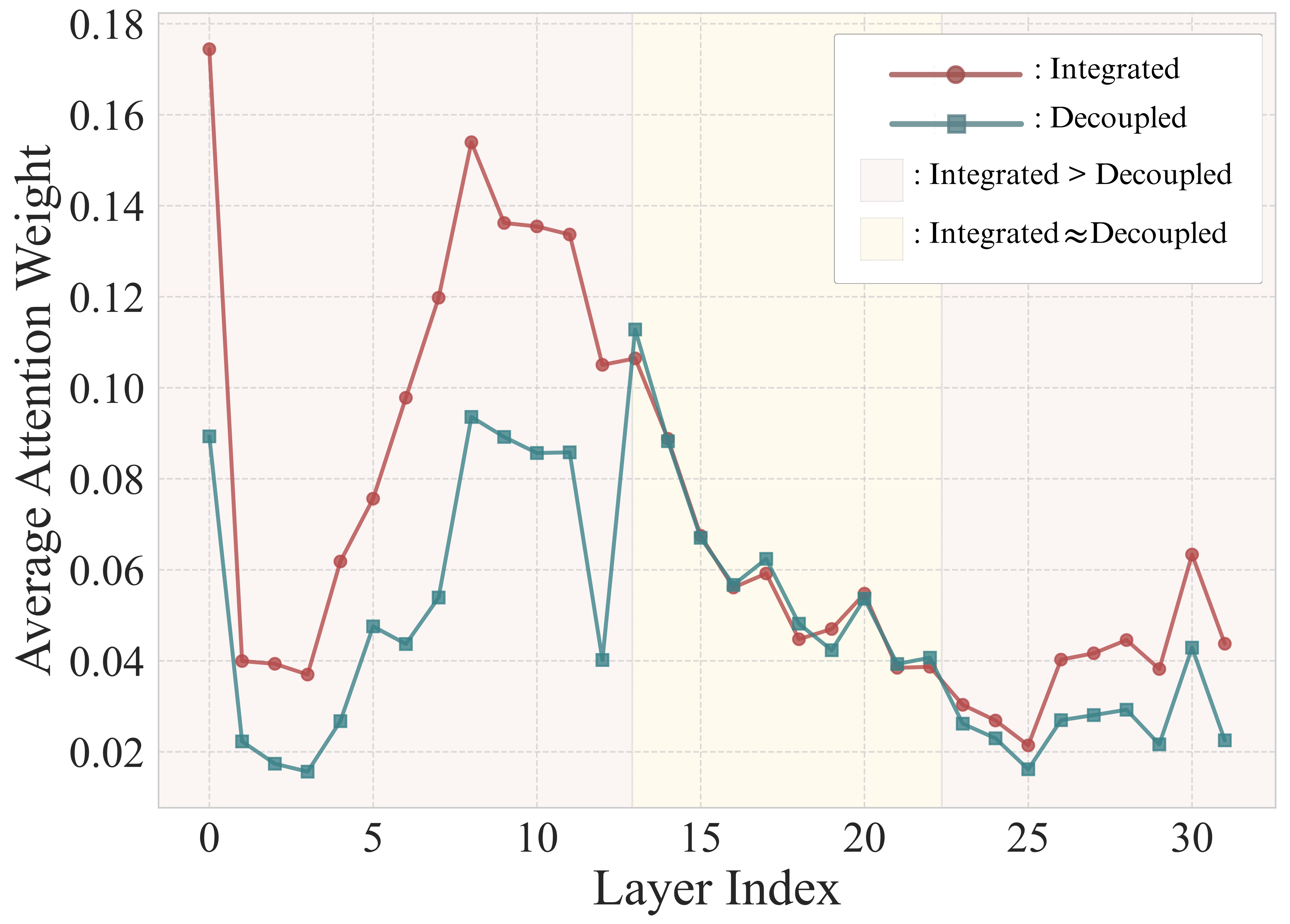}%
    \caption{\textbf{Layer-wise average attention to self-generated examples.}
    The x-axis represents layer indices, and the y-axis shows the average attention that tokens generated during test-question solving assign to self-generated example tokens.
    On LLaMA 3.1 8B Instruct, \textit{Integrated} exhibits higher attention scores than \textit{Decoupled} in the lower layers (0–12) and upper layers (23–31) among the 32 layers (0–31).}
    \label{fig:attention_layer_q}
    \vspace{-5mm}
\end{figure}

\subsubsection{Multi-shot Settings}
\label{subsubsec:multi_shot}
Although our primary experiments focus on zero-shot and one-shot prompting, we also tested 2-shot and 4-shot configurations to determine if Integrated retains its advantage over Decoupled with more in-context examples. 
Table~\ref{tab:multi_shot_llama33_gsm8k} shows that \textbf{Integrated} still outperforms \textbf{Decoupled} under multi-shot conditions.
Interestingly, in the Integrated setup, moving from 1-shot to 2-shot yields a noticeable gain (35.19\% to 38.93\%), and 4-shot performance (38.24\%) remains higher than 1-shot as well. 
By contrast, in the Decoupled setup, adding more in-context examples does not improve results: 1-shot achieves 33.59\%, whereas 2-shot and 4-shot drop to 31.91\% and 32.44\%, respectively.
These findings suggest that the benefits of Integrated prompting are not restricted to minimal-context scenarios, persisting even as the number of exemplars increases.

\subsubsection{Attention Distribution Analysis}
\label{subsubsec:attention_analysis}
To better understand why the Integrated prompting approach outperforms Decoupled prompting, we conducted an in-depth examination of how the model distributes its internal attention during the generation process. Specifically, we focus on the token-generation phase, where the model produces its final answer to the test question. At this stage, the model has either generated (Integrated) or read (Decoupled) the self-generated content. We use the open-source \textsc{LLaMA-3.1-8B-Instruct} model and measure how each newly generated query token (\textit{i.e.}, every token produced while solving the test question) attends to two critical regions of the input prompt: (1) \textbf{self-generated example tokens} and (2) \textbf{test question tokens}.

\noindent Our analysis comprised a multi-step approach designed to precisely characterize the internal attention mechanisms:
\begin{itemize}
    \item \textbf{Identify each query token.} 
    Let the number of tokens the model generates while producing the answer to the test question be $T$. Then, every $T$ token is treated as a query token.
    
    \item \textbf{Compute its attention to each region.} 
    For each query token, we analyze the attention assigned to the test-question tokens and to the self-generated-example tokens. We aggregate these attention weights as a single average value per region across all heads and layers. (\(T \times \text{Heads} \times \text{Layers} \rightarrow T\))
    
    \item \textbf{Aggregate across all solution tokens.} 
    We repeat the above step for every token in the final answer, thereby collecting attention values for the test-question region and the self-generated-example region within each sample. (2 sets of size \(T\), one for each region)
    
    \item \textbf{Obtain final mean scores per sample.} 
    We average across all query tokens in that answer to yield one mean attention score for the test-question region and another for the self-generated-example region. Each test sample thus produces a pair of mean scores, \textit{i.e.}, one for each region. (2 sets of size \(T \rightarrow\) 2 mean scores)
\end{itemize}

Figure~\ref{fig:attention_distribution_test} shows the distributions of mean attention towards (a) the test question tokens and (b) the self-generated example tokens, comparing Integrated (red) and Decoupled (green) prompting. From these results, we identify two key differences in attention allocation:

\paragraph{(1) Integrated allocates more attention to self-generated examples.}
Under Integrated prompting, the model consistently assigns higher attention to the self-generated example tokens compared to Decoupled prompting (paired t-test, $p < 10^{-10}$).

\paragraph{(2) Decoupled devotes more attention to the test question.}
Decoupled prompting leads to significantly greater attention towards test-question tokens, highlighting a contrasting internal emphasis compared to Integrated (paired t-test, $p < 10^{-10}$).

\subsubsection{Layer-wise Attention Pattern Analysis}

To gain deeper insights into the differences between Integrated and Decoupled approaches, we conducted a comprehensive analysis of layer-wise attention patterns across all 32 layers (0-31) of the LLaMA-3.1-8B-Instruct model. As illustrated in Figure~\ref{fig:attention_layer_q}, attention to self-generated examples varies significantly across different layers during test question solving. By aggregating attention values across all test samples and tokens generated during solution phases, we identified distinct attention behaviors at various model depths.

Our results align with established findings regarding Transformer architectures. Lower layers (approximately layers 0–12) primarily encode lexical and syntactic features of inputs \citep{bert2019tenny, jawahar2019what}, and Integrated prompting shows notably higher attention scores in these layers. This suggests that Integrated prompting effectively utilizes initial contextual grounding, embedding the self-generated examples actively into the model’s representations. Previous analyses indicate that early attention heads often specialize in capturing token-level and structural information \citep{voita2019analyzing, clark2019what}, supporting our observation that early contextual embedding is crucial in Integrated prompting.

Interestingly, attention differences diminish in middle layers (layers 13–22), potentially reflecting a transitional encoding phase. These layers have been identified to host attention heads functioning as "induction heads," responsible for recognizing and aligning relevant contextual information from prompts \citep{olsson2022head, wang2023label}. Given Decoupled prompting’s structure of providing similar examples without their generative context, these induction mechanisms likely activate similarly across both prompting strategies, explaining the minimal differences observed at this intermediate stage.

In contrast, significant attention differences re-emerge in upper layers (layers 23–31), aligning well with recent studies highlighting that deeper Transformer layers are specialized for task-specific reasoning and final decision-making \citep{sajjad2022analyzing, sia2024where}. Here, the self-generated examples in Integrated prompting seem to act as strong, readily accessible reference points, substantially guiding the reasoning process and final answer generation. Studies have indicated upper-layer attention patterns in Transformers heavily influence the integration of contextual knowledge for accurate predictions \citep{kovaleva2019revealing, abnar2020quantifying}. This observation explains why Integrated prompting consistently outperforms Decoupled prompting: Integrated method makes the self-generated examples easily accessible to the model at deeper layers, where critical reasoning and final decisions occur.

Overall, our findings suggest that the advantage of Integrated prompting arises from a dual-layered mechanism: stronger early-layer encoding of contextual information and more effective utilization of this context in later reasoning layers. These observations reinforce prior findings regarding the differentiated roles of Transformer layers \citep{michel2019sixteen, wang2024grokking} and emphasize the significance of explicitly incorporating example-generation processes into prompting strategies to maximize model reasoning effectiveness.

\section{Discussion}
\label{sec:discussion}
Having shown that \textbf{Integrated} prompting consistently outperforms \textbf{Decoupled}, we explore additional experiments and analyses to better understand the underlying reasons for this advantage. Specifically, we address the following points:

\begin{enumerate}[leftmargin=*,noitemsep,topsep=2pt]
\item Is the improvement simply due to generating additional text?
\item Does explicitly repeating examples without the generation process affect performance?
\item Could differences in example quality between Integrated and Decoupled explain the gap?
\item Does the advantage persist with Chain-of-Thought prompting?
\item How does explicitly stating intentions behind example generation influence performance?
\item Is the advantage consistent across different problem categories and difficulty levels?
\end{enumerate}

We discuss each of these questions in detail in the subsequent subsections. 

\subsection{Dummy Examples: Is Performance Gain Just About Text Length?}
\label{subsec:dummy}
Another question is whether performance gains arise purely from generating extra text. To explore this, we replaced the self-generated math problem in Integrated with 50 random English characters (“dummy” content). Surprisingly, even this “dummy Integrated” setup performed slightly better than Zero-shot or Decoupled, though it fell short of the original Integrated. For example, with \textsc{LLaMA-3.3-70B-Instruct} on \textsc{GSM8K}, the dummy-Integrated configuration reached 33.82\%, compared to roughly 32\% for both Zero-shot and Decoupled. This partial improvement aligns with prior findings that simply making the model generate some intermediate text can help it engage more actively in the solving process. However, genuinely relevant content still yields the highest accuracy.

\subsection{The Effect of Example Location vs. Generation Process}
\label{subsec:example_location}

To further isolate whether the location of examples or the generation process drives performance improvements, we designed an experiment where models explicitly repeat the provided Decoupled example before solving.

\begin{table}[t]
\centering
\small
\begin{tabular}{lccc}
\toprule
\textbf{Model} & \textbf{Integrated} & \textbf{Decoupled} & \textbf{Repeat} \\
\midrule
\gptfouro & \textbf{37.86} & 32.65 & 35.93 \\
\textsc{Gemini-1.5} & \textbf{38.40} & 36.35 & 37.78 \\
\bottomrule
\end{tabular}
\vspace{-1.5mm}
\caption{
\textbf{Accuracy(\%) on \textsc{MATH} comparing Integrated, Decoupled, and Repeat prompting.}
\textbf{Repeat} denotes explicitly repeating the Decoupled example within the prompt before solving the question.
\vspace{-5.5mm}
}
\label{tab:repeat_results}
\end{table}

As shown in Table~\ref{tab:repeat_results}, the "Repeat" approach outperforms the original Decoupled method but falls short of Integrated prompting. These results suggest that while example location contributes to performance, the integrated generation process remains the principal factor driving improvement.

\subsection{Quality of Generated Examples}
\label{subsec:quality_vs_process}
A natural concern is whether Integrated simply produces better-quality examples, thus explaining the performance gap. To check this, we directly reused identical examples generated under Integrated prompting in a Decoupled prompt. The resulting accuracies were 52.16\% on \textsc{GSM8K} and 33.68\% on \textsc{MATH} (using GPT-4o), which remained lower than Integrated performance. This confirms that the performance gain is driven not by example quality alone, but by the structured reasoning process in Integrated prompting.

\subsection{Effectiveness with Chain-of-Thought Reasoning}
\label{subsec:cot_reasoning}

Our main experiments employ a non-CoT setting to isolate the effect of self-generation without the confounding factor of step-by-step reasoning. However, it's important to verify whether our findings extend to Chain-of-Thought (CoT) settings as well.

\begin{table}[t]
\centering
\begin{tabular}{lccc}
\toprule
\textbf{Model} & \textbf{Zero-shot} & \textbf{Integrated} & \textbf{Decoupled} \\
\midrule
\textsc{GPT} & 75.36 & \textbf{76.59} & 72.90 \\
\textsc{Gemini} & 78.23 & \textbf{79.06} & 78.65 \\
\bottomrule
\end{tabular}
\vspace{-1mm}
\caption{
\textbf{Accuracy of GPT-4o-mini and Gemini-1.5-Flash with CoT prompting on \textsc{MATH}.}
Even with CoT reasoning applied, Integrated maintains its advantage.
}
\vspace{-4mm}
\label{tab:cot_results}
\end{table}

As shown in Table~\ref{tab:cot_results}, Integrated prompting maintains its performance advantage even with CoT reasoning, though with a narrower gap. This aligns with findings from \citet{li2024are} and \citet{yasunaga2024large}, who similarly report that self-generated examples consistently enhance performance regardless of whether CoT is employed. The smaller performance difference suggests that explicit reasoning may partially compensate for the benefits of the generation process, but the core advantage of Integrated prompting persists across reasoning frameworks.

\subsection{The Role of Intention in Problem Generation}
\label{subsec:intention}

One might conjecture that if LLMs explicitly stated their reasoning or rationale for generating each example, the gap between  Integrated and Decoupled could disappear. To test this, we modified the prompting strategies so that the model not only generated each example but also explained why it created that particular problem (e.g., “\emph{I created this problem to illustrate a tricky geometry concept}”). Even with these additional self-generated ‘intentions,’ \textbf{Integrated} remained superior. For instance, as shown in Table~\ref{tab:intention_results}, \textit{Decoupled + intention} achieved 41.60\% while \textit{Integrated + intention} reached 45.50\% with \llamathree. 
Interestingly, both intention-augmented methods outperformed their respective baselines without intentions, suggesting that clarifying why a problem was generated can be beneficial. 
This points to a potential future direction for LLM-based synthetic data generation or self-generation techniques.

\begin{figure}[t] 
    \centering
    \includegraphics[width=\columnwidth]{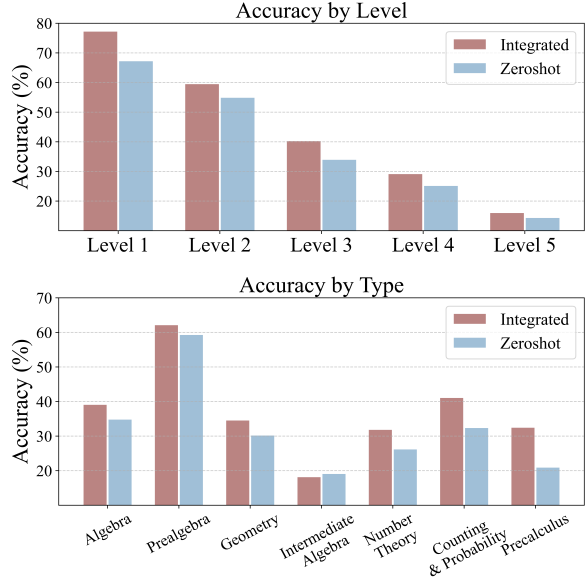} 
    \vspace{-3mm}
    \caption{
    \textbf{Performance across problem categories and difficulty levels on MATH dataset.}
    (Top) Accuracy comparison between Zero-shot prompting and 
    Integrated prompting across different problem categories.
    (Bottom) The same comparison across difficulty levels.
    }
    \label{fig:additional_result_2}
    \vspace{-1mm}
\end{figure}

\begin{table}[t]
\centering
\begin{tabular}{lcc}
\toprule
\textbf{Setup} & \textbf{No-intention} & \textbf{+Intention} \\
\midrule
\textbf{Decoupled} & 33.59 & 41.60 \\
\textbf{Integrated} & \textbf{35.19} & \textbf{45.50} \\
\bottomrule
\end{tabular}
\caption{
\textbf{Effect of explicit ``intention'' statements on \textbf{GSM8K}.}
We compare Decoupled vs.\ Integrated with and without intentions. 
}
\vspace{-5mm}
\label{tab:intention_results}
\end{table}

\subsection{Effects across Domains and Difficulty Levels on MATH dataset}
\label{subsec:comparison_dec_int}
We assessed whether Integrated excels only in certain mathematical subfields or problem difficulty levels.
As illustrated in Figure \ref{fig:additional_result_2}, the performance gains were consistent rather than exploiting any particular category or subset of problems.

\section{Conclusion}
In this paper, we investigated the role of self-generation prompts in enhancing the reasoning capabilities of LLMs. Through experiments on \textsc{MATH} and \textsc{GSM8K} datasets, we demonstrated that Integrated prompting achieves substantial accuracy gains, significantly outperforming other prompting strategies. Importantly, our findings reveal that these improvements arise not from the generated examples themselves but from the process of creating them. This emphasizes the need for a deeper understanding of the problem-creation process and highlights the potential of self-generation as a powerful strategy for enhancing the reasoning capabilities of LLMs. As a future direction, we plan to investigate how iterative refinement, self-verification, or external knowledge retrieval can further strengthen the creation-and-solving process, which remains a promising avenue for building more robust and interpretable LLM-based systems.

\section*{Limitations}
While our findings underscore the importance of the self-generation process in improving LLM reasoning, our study focuses primarily on mathematical problem solving (\textsc{MATH} and \textsc{GSM8K}). Although math tasks offer a rigorous test bed for reasoning, future work could examine whether the same benefits extend to other domains (e.g., legal or biomedical texts) that require fundamentally different forms of knowledge and context. 

In addition, our experiments were conducted using five LLM architectures. While these span a range of model sizes, further investigation with additional model families and scales may help to understand how architectural nuances influence the efficacy of self-generation. Our attention analysis, though informative, was also restricted to a single open-source model. Exploring attention or other interpretability techniques across multiple architectures could provide deeper insight into the underlying mechanisms.

\section*{Ethical Considerations}
This work focuses on improving the reasoning capabilities of LLMs through self-generation prompts, with experiments conducted on publicly available mathematical reasoning datasets. While our research does not directly involve sensitive data or applications with significant consequences, we recognize the broader implications of deploying LLMs in real-world scenarios. Models enhanced through self-generation techniques may still generate incorrect or misleading outputs, particularly in domains requiring precise or factual reasoning. To mitigate potential misuse, we emphasize the importance of human oversight when applying these methods in critical domains. 

\bibliography{custom}

\newpage

\appendix

\begin{figure*}[h!]
    \centering
    \fbox{
        \begin{minipage}{0.95\textwidth}
        \textbf{(a) Zero-shot Prompt Example} \\
        \textbf{Prompt:} \\
        Solve the following math word problems. You should only provide the answer that goes into the part [YOUR ANSWER HERE] without any solution.\\
        
        Test Question: How many ways are there to put 5 balls in 3 boxes if the balls are distinguishable and the boxes are distinguishable?\\
        Answer of Test Question: [YOUR ANSWER HERE]\\
        
        \textbf{Generated Response:} \\
        243
        \end{minipage}
    }
    \vspace{3mm}
    \fbox{
        \begin{minipage}{0.95\textwidth}
        \textbf{(b) Integrated Prompt Example} \\
        \textbf{Prompt:} \\   
        Step 1: Take the role of a problem creator and carefully analyze the given Test Question. Based on this analysis: \\
        - Create one similar problem that shares the key concepts, structure, or logic with the Test Question. Ensure that the created problem is directly helpful for solving the Test Question. \\
        - Provide only the question and correct final answer to the created problem. \\

        Important Guidelines for the Created Problem: \\        
        - Avoid generating problems that are significantly different or unrelated to the Test Question. \\
        - Keep the complexity of the created problem comparable to that of the Test Question. \\
        - It should help reinforce the skills required to solve the Test Question. \\

        Test Question: How many ways are there to put 5 balls in 3 boxes if the balls are distinguishable and the boxes are distinguishable? \\
        
        <Result of Step 1> \\
        Created Problem: [YOUR ANSWER HERE] \\
        Answer of Created Problem: [YOUR ANSWER HERE] \\

        Step 2: Solve the following math word problems. You should only provide the answer that goes into the part [YOUR ANSWER HERE] without any solution.\\

        Solve the given Test Question by using the created problem in Step 1 to guide your answer to the Test Question. \\

        <Result of Step 2> \\
        Answer of Test Question: [YOUR ANSWER HERE] \\
    
        \textbf{Generated Response:} \\
        Step 1:\\ 
        Created Problem: How many ways are there to put 4 balls in 2 boxes if the balls are distinguishable and the boxes are distinguishable?\\ 
        Answer of Created Problem: 16 \\

        Step 2:\\
        Answer of Test Question: 243
        \end{minipage}
    }
    \caption{Prompt examples for Zero-shot, and Integrated prompting. Each example illustrates the input-output structure of prompting.}
    \label{fig:prompt-examples}
\end{figure*}
\begin{figure*}[h!]
    \centering
    \fbox{
        \begin{minipage}{0.95\textwidth}
        \textbf{(c) Decoupled Prompt Example} \\
        \textbf{Prompt 1:} \\   
        Take the role of a problem creator and carefully analyze the given Test Question. Based on this analysis: \\
        - Create one similar problem that shares the key concepts, structure, or logic with the Test Question. Ensure that the created problem is directly helpful for solving the Test Question. \\
        - Provide only the question and correct final answer to the created problem. \\
    
        Important Guidelines for the Created Problem: \\
        - Avoid generating problems that are significantly different or unrelated to the Test Question. \\
        - Keep the complexity of the created problem comparable to that of the Test Question. \\
        - It should help reinforce the skills required to solve the Test Question. \\
    
        Test Question: How many ways are there to put 5 balls in 3 boxes if the balls are distinguishable and the boxes are distinguishable? \\
    
        <Result> \\
        Created Problem: [YOUR ANSWER HERE] \\
        Answer of Created Problem: [YOUR ANSWER HERE] \\
        
        \textbf{Generated Response:} \\
        Created Problem: How many ways are there to put 4 balls in 2 boxes if the balls are distinguishable and the boxes are distinguishable? \\
        Answer of Created Problem: 16 \\
        
        \textbf{Prompt 2:} \\
        Solve the following math word problems. You should only provide the answer that goes into the part [YOUR ANSWER HERE] without any solution. \\
        
        Solve the given Test Question by using the One-shot Example provided below to guide your answer to the Test Question. \\
        
        One-Shot Example: \\
        Example Question: How many ways are there to put 4 balls in 2 boxes if the balls are distinguishable and the boxes are distinguishable? \\
        Example Answer: 16 \\
        
        Test Question: How many ways are there to put 5 balls in 3 boxes if the balls are distinguishable and the boxes are distinguishable? \\
        
        <Result> \\
        Answer of Test Question: [YOUR ANSWER HERE] \\
        
        \textbf{Generated Response:} \\
        Answer of Test Question: 243
        \end{minipage}
    }
    \caption{Prompt examples for Decoupled prompting. Each example illustrates the input-output structure of prompting.}
    \label{fig:prompt-examples_2}
\end{figure*}

\section{Detailed Methodology}
\label{sec:method}

This section provides a comprehensive overview of the methodology employed in our experiments, addressing the problem definition, prompting strategies, and experimental setup in detail.

\subsection{Problem Definition}
The primary goal of this study is to investigate the underlying reasons why self-generated examples improve reasoning performance in Large Language Models (LLMs). Formally, we define our research problem as follows:

Given a test problem $q_{test}$, our objective is to evaluate how different methods of generating and utilizing self-generated examples affect the performance of an LLM in solving $q_{test}$. We compare performance under three distinct prompting strategies, analyzing whether observed improvements arise primarily from the process of generating examples or merely from the generated examples themselves.

\subsection{Prompting Strategies}
We systematically compare three prompting methods designed to isolate the impact of the self-generation process:

\paragraph{Zero-shot Prompting.} The simplest approach, which involves presenting the LLM with a test question $q_{test}$ directly without additional context or examples:
\begin{itemize}
\item Prompt Structure: \textit{"Solve the following problem: $q_{test}$."}
\end{itemize}

\paragraph{Integrated Prompting.} This approach integrates problem creation and solving in a single prompt. The LLM first generates a related problem $q_{gen}$ and its solution $a_{gen}$ before solving $q_{test}$. This method aims to engage the LLM actively in a unified process of creation and solving:
\begin{itemize}
\item Prompt Structure: \textit{"First, create and solve a related problem. Then, using that as a reference, solve the following problem: $q_{test}$."}
\end{itemize}

\paragraph{Decoupled Prompting.} This method decouples problem creation from problem-solving. A self-generated example $(q_{gen}, a_{gen})$ is generated beforehand (either externally or in a previous interaction), and provided directly to the LLM as a static context for solving $q_{test}$. This setup assesses the utility of examples alone, without the context of their creation:
\begin{itemize}
\item Prompt Structure: \textit{"Here is an example: Problem $q_{gen}$, Answer $a_{gen}$. Now, solve the following problem: $q_{test}$."}
\end{itemize}

These three distinct approaches allow us to rigorously examine the specific contributions of the problem-generation process versus the mere presence of self-generated examples in enhancing LLM reasoning.

\begin{figure}[t] 
    \centering
    \includegraphics[width=\columnwidth]{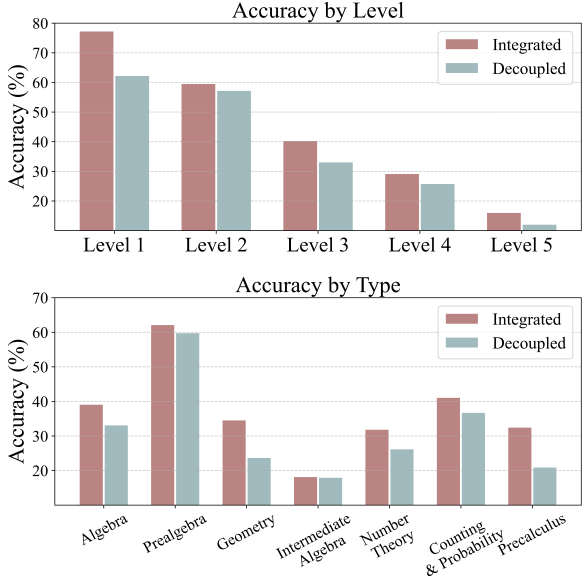} 

    \caption{
    \textbf{Performance across problem categories and difficulty levels on MATH dataset.}
    (Top) Accuracy comparison between Decoupled prompting and 
    Integrated prompting across different problem categories.
    (Bottom) The same comparison across difficulty levels.
    }
    \label{fig:additional_result_2}
\end{figure}

\section{Examples of Prompts and Generated Responses}
\label{sec:prompt_examples}
This section provides examples of prompts and generated responses used in our experiments for each prompting strategy. We include Zero-shot, Integrated, and Decoupled prompting examples to clarify the input-output structure and differences across methods. 

As shown in Figure \ref{fig:prompt-examples} and \ref{fig:prompt-examples_2}, each strategy demonstrates distinct ways in which the model interacts with the provided task. Zero-shot prompting directly solves the test question without additional context. Integrated prompting combines problem creation and solving within a single prompt, providing a unified reasoning process. In contrast, Decoupled prompting separates problem creation from solving, where generated examples are reused as in-context guidance but without the reasoning context from their creation.

\section{Compliance with Data Usage Policies}

\subsection{Dataset Citation and Usage}
Our experiments utilized the publicly available \textsc{MATH} \citep{hendrycksmath2021} and \textsc{GSM8K} \citep{cobbe2021gsm8k} datasets, which are widely used benchmarks for evaluating reasoning capabilities of LLMs. Both datasets were used in accordance with their intended purpose of assessing model performance on mathematical problem solving.

\subsection{License and Terms of Use}
The \textsc{MATH} and \textsc{GSM8K} datasets are released for academic and research purposes. We adhered to the usage policies specified for each dataset.

\subsection{Data Anonymization}
Both \textsc{MATH} and \textsc{GSM8K} datasets consist of mathematical problems and solutions, and do not contain any personally identifiable information.

\section{Use of AI Tools}
AI tools, including \gptfouro, were utilized in the experiments of our study. Additionally, AI-assisted writing tools, such as ChatGPT, were employed to refine the paper’s narrative and ensure clarity. All AI tools were used responsibly, with oversight to maintain the integrity and originality of the research.

\newpage
\section*{Summary of Changes (May 2025 Revision)}

We sincerely thank the meta-reviewer and reviewers for their valuable feedback, which significantly improved our paper. Below, we summarize key revisions made in response to reviewer suggestions from the ACL ARR 2025 February submission:

\begin{enumerate}

\item \textbf{Expanded Experimental Analysis}
\begin{itemize}
\item \textit{New Models:} Included additional results using GPT-4.1, extending our previous evaluation to five representative LLM architectures.
\item \textit{Chain-of-Thought (CoT) Evaluation:} Added CoT prompting results for GPT-4o-mini and Gemini-1.5-Flash, demonstrating that Integrated prompting's advantage persists under CoT reasoning.
\end{itemize}

\item \textbf{Improved Attention Analysis}
\begin{itemize}
\item \textit{Layer-wise Analysis:} Revised our attention analysis to clearly explain attention patterns across transformer layers, highlighting key differences between Integrated and Decoupled prompting at early (contextual encoding) and later (reasoning and decision-making) layers.
\item \textit{Enhanced Visualization:} Updated figures to better illustrate these layer-specific differences.
\end{itemize}

\item \textbf{Additional Experiments in Discussion Section}
\begin{itemize}
\item \textit{Example Repetition Experiment:} Implemented a new experiment (suggested by Reviewer U4tT) where models explicitly repeat provided examples before solving, clarifying that the improvement primarily arises from the integrated generation process rather than example location alone.
\item \textit{Explicit Intention Prompts:} Extended experiments with explicit ``intention'' statements, further reinforcing the advantage of Integrated prompting.
\end{itemize}

\item \textbf{Writing and Structural Enhancements}
\begin{itemize}
\item \textit{Detailed Methodology:} Added a comprehensive methodology section in the appendix to clearly define experimental procedures and prompting strategies, addressing concerns about methodological transparency.
\item \textit{Expanded Discussion:} Substantially enriched the discussion with detailed analyses, clearer experimental results, and broader implications to meet the depth expected of a long paper format.
\end{itemize}

\end{enumerate}

These changes directly respond to the reviewers' comments, clarifying our main findings and enhancing the overall quality and comprehensiveness of the manuscript.
 
\end{document}